\documentclass[pdflatex,sn-apa]{sn-jnl}

\usepackage{graphicx}%
\usepackage{multirow}%
\usepackage{amsmath,amssymb,amsfonts}%
\usepackage{amsthm}%
\usepackage{mathrsfs}%
\usepackage[title]{appendix}%
\usepackage{xcolor}%
\usepackage{textcomp}%
\usepackage{manyfoot}%
\usepackage{booktabs}%
\usepackage{algorithm}%
\usepackage{algorithmicx}%
\usepackage{algpseudocode}%
\usepackage{listings}%
\usepackage{adjustbox}
\usepackage{float}

\theoremstyle{thmstyleone}%
\theoremstyle{thmstyletwo}%

\theoremstyle{thmstylethree}%

\begin{document}

\title[Article Title]{A Robust Deep Learning System for Motor Bearing Fault Detection: Leveraging Multiple Learning Strategies and a Novel Double Loss Function}


\author*[1]{\fnm{Khoa} \sur{Tran}}\email{tdkhoa@aiware.website}
\author[2]{\fnm{Lam} \sur{Pham}}
\author*[3]{\fnm{Vy-Rin} \sur{Nguyen}}\email{RinNV@fe.edu.vn}
\author[4]{\fnm{Ho-Si-Hung} \sur{Nguyen}}

\affil*[1]{\orgname{AIWARE Limited Company}, \orgaddress{\city{Da Nang}, \postcode{550000}, \country{Vietnam}}}

\affil[2]{\orgname{AIT Austrian Institute of Technology GmbH}, \orgaddress{\street{Giefinggasse 4}, \city{Vienna}, \postcode{1210}, \country{Austria}}}

\affil[3]{\orgdiv{Software Engineering Department}, \orgname{FPT University}, \orgaddress{\city{Da Nang}, \postcode{550000}, \country{Vietnam}}}

\affil[4]{\orgdiv{The University of Danang - University of Science and Technology}, \orgaddress{\street{54 Nguyen Luong Bang, Hoa Khanh Bac, Lien Chieu}, \city{Da Nang}, \postcode{550000}, \country{Vietnam}}}






\abstract{Motor bearing fault detection (MBFD) is critical for maintaining the reliability and operational efficiency of industrial machinery. Early detection of bearing faults can prevent system failures, reduce operational downtime, and lower maintenance costs. In this paper, we propose a robust deep learning-based system for MBFD that incorporates multiple training strategies, including supervised, semi-supervised, and unsupervised learning. To enhance the detection performance, we introduce a novel double loss function. Our approach is evaluated using benchmark datasets from the American Society for Mechanical Failure Prevention Technology (MFPT), Case Western Reserve University Bearing Center (CWRU), and Paderborn University's Condition Monitoring of Bearing Damage in Electromechanical Drive Systems (PU). Results demonstrate that deep learning models outperform traditional machine learning techniques, with our novel system achieving superior accuracy across all datasets. These findings highlight the potential of our approach for practical MBFD applications.}




\maketitle

\section{Introduction}
\label{intro}
Rolling bearing failures account for a significant proportion, ranging from 40\% to 70\%, of breakdowns in electro-mechanical drive systems and motor installations \citep{bonnett, djeddi}. These failures result in considerable maintenance costs and operational downtime for motor-based systems. To address this, there is a growing need for maintenance strategies capable of continuous monitoring and automatic detection of motor bearing faults. Vibration data, collected via acceleration sensors integrated into motors, is widely used for fault detection in rolling bearings. Recently, machine learning and deep learning models have gained popularity in the analysis of vibration data for bearing fault detection, achieving promising results \citep{tong2025modulated, zhang2023enhancement, swami2024enhanced, ceylan2024cost, zhao2024improved}.

Traditional machine learning-based approaches to motor bearing fault detection (MBFD) rely on extracting features from both time and frequency domains. Time-domain features such as mean, root mean square (RMS), kurtosis, peak-to-peak, variance (Var), standard deviation (SD), and shape factor are commonly used~\citep{goyal2017condition, omoregbee2018diagnosis}. Frequency-domain features like root mean square frequency (RMSF), center frequency (CF), and variance of frequency (VF) are also extracted~\citep{freq_dom_01, freq_dom_02}. These features are combined into a multi-domain representation for classification. However, not all features are equally significant, and several methods have been proposed to optimize feature selection. Techniques like particle swarm optimization (PSO)~\cite{dixit2023bmudf} and principal component analysis (PCA)~\cite{wang2004data, verma2023fcmcps} have been employed to reduce the feature set to improve model accuracy. Furthermore, entropy-based approaches have been introduced to capture the uncertainty of features, enhancing their robustness~\citep{ent_dom_01, ent_dom_02}.

While machine learning-based methods have proven effective, deep learning-based approaches have gained traction due to their ability to learn feature representations directly from raw vibration signals. Convolutional Neural Networks (CNNs), Recurrent Neural Networks (RNNs), autoencoders, and hybrid CNN-RNN architectures have been successfully applied for MBFD~\citep{zhang2024multi, sun2017intelligent, li2021speckle, ahakonye2024low, li2020novel, mao2019imbalanced}. CNNs, in particular, have been widely adopted due to their ability to capture spatial and temporal patterns in vibration data without requiring hand-crafted features. Moreover, transfer learning techniques have been employed to enhance performance further. For instance, pre-trained ResNet-50 models have been fine-tuned for bearing fault detection tasks, and transfer learning has also been applied to multilayer perceptron architectures~\citep{transfer_01, transfer_02}.

In this paper, we address several key gaps in the existing literature:

\begin{enumerate}

\item While machine learning and deep learning-based methods have been explored independently for MBFD, there is a lack of a comprehensive comparison between the two approaches. We aim to fill this gap by comparing both approaches on multiple datasets.

\item For the specific problem of MBFD, we propose a novel deep learning system called Robust-MBFD, which integrates supervised, semi-supervised, and unsupervised learning strategies. This allows us to explore a broader range of training strategies and address the challenges of limited labeled data.

\item To enhance the training of the deep learning system, we propose a novel double loss function that combines Triplet loss and Center loss, addressing the issue of intra-class variance and improving the model's ability to generalize across datasets.

\item Finally, we conduct extensive experiments using benchmark datasets, including the American Society for Mechanical Failure Prevention Technology (MFPT)~\citep{mfpt_data}, Case Western Reserve University Bearing Center (CWRU)~\citep{cwru_data}, and Paderborn University's Condition Monitoring of Bearing Damage in Electromechanical Drive Systems (PU)~\citep{pu_data}. We establish standard training-testing splits to facilitate fair comparisons between proposed systems, demonstrating the robustness and generalization capabilities of the Robust-MBFD system across multiple datasets.
\end{enumerate}

This work provides a comprehensive analysis of machine learning and deep learning-based MBFD systems, presenting novel contributions that advance the state-of-the-art in fault detection.


\section{Machine Learning and Deep Learning-Based Systems for Bearing Fault Detection}
In this paper, we propose two approaches to develop a system for MBFD, which are based on (1) machine learning-based and (2) deep learning-based techniques.

\subsection{Proposed Machine Learning-Based Systems}
\label{ml_sys}

\begin{figure}[h]
    \centering
    \includegraphics[width =1.0\linewidth]{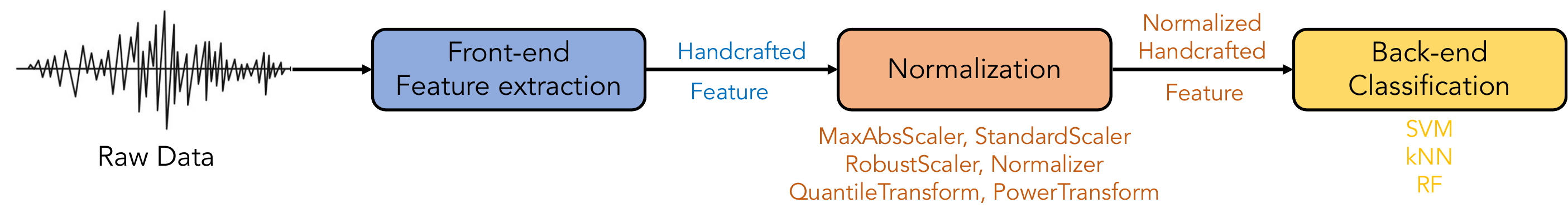}
    \caption{High-level architecture of the proposed machine learning-based systems.}
    \label{fig:ml_fig}
\end{figure}

The high-level architecture of our proposed machine learning-based systems is illustrated in Fig.~\ref{fig:ml_fig}.
As shown, the systems consist of three main components: front-end feature extraction, normalization, and back-end classification.

\textbf{Front-end feature extraction:} 
A total of 16 features are selected, comprising 8 time-domain features and 8 frequency-domain features. The time-domain features include root mean square, variance, peak values, kurtosis, skewness, peak-to-peak value, line integral, and four-factor values, as inspired by \cite{kimotho2014approach}. The frequency-domain features consist of spectral centroid, spectral bandwidth, spectral flatness, and roll-off frequency, as inspired by \cite{klapuri2007signal}. These 16 features are concatenated to form hand-crafted features representing the input vibration data.

\textbf{Normalization:} 
The hand-crafted features undergo normalization using various techniques, such as max absolute scaler (MAS), standard scaler (SS), robust scaler (RS), normalization (Norm), quantile transformer (QT), and power transformer (PT) methods, as proposed by \cite{pires2020homogeneous}.

\textbf{Back-end classification:} 
The normalized features are classified using Support Vector Machine (SVM), k-nearest neighbors (kNN), and Random Forest (RF) models. Specific model parameters, optimized through extensive experimentation, are detailed in Table~\ref{table:set_model}.

\begin{table}[h]
    \caption{Back-end machine learning-based classification models and their parameter settings.}
    \centering
    \begin{tabular}{cl}
        \toprule
        \textbf{Back-end classification Models} & \textbf{Setting Parameters} \\
        \midrule
        SVM &  \begin{tabular}[t]{@{}l@{}}
                                            $C=1.0$ \\
                                            kernel=`RBF' \\
                                            gamma=`scale'
                                        \end{tabular} \\
        \midrule
        kNN & \begin{tabular}[t]{@{}l@{}}
                                        $n\_neighbors = 5$ \\
                                        weights = 'uniform' \\
                                        leaf\_size = 30 \\
                                        $p=2$
                                    \end{tabular} \\
        \midrule
        RF & \begin{tabular}[t]{@{}l@{}}
                                    max depth of tree = 20, \\
                                    number of trees = 100
                                \end{tabular} \\
        \bottomrule
    \end{tabular}
    \label{table:set_model}
\end{table}

\subsection{Proposed Deep Learning-based Systems}
\label{dl_sys}

\begin{figure}[h]
    \centering
    \includegraphics[width =1.0\linewidth]{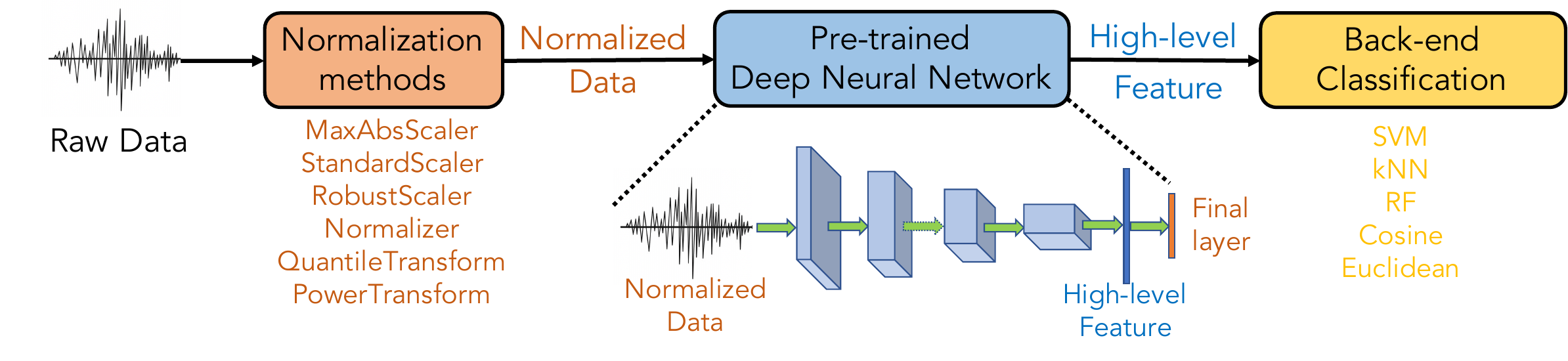}\\
	\caption{High-level architecture of the proposed deep learning-based systems.}
    \label{fig:dl_fig}
\end{figure}

The high-level architecture of our proposed deep learning-based systems is illustrated in Fig.~\ref{fig:dl_fig}. It consists of three main components: normalization, high-level feature extraction, and back-end classification.

\textbf{Normalization:} Raw vibration data is normalized using methods adapted from the machine learning-based systems discussed in Section~\ref{ml_sys}.

\textbf{High-level feature extraction:} The normalized vibration data undergoes training in a deep neural network. After training, the data is passed through the pre-trained network to extract feature maps, which represent the output of the final fully connected layer (as shown in Fig.~\ref{fig:dl_fig}). These feature maps are termed high-level features.

\textbf{Back-end classification:} High-level features are classified into specific classes using machine learning models such as SVM, kNN, and RF mentioned in Section~\ref{ml_sys}. Additionally, Euclidean and Cosine distance measurement methods are employed to provide alternative classification approaches based on similarity metrics.

\textbf{Why not use deep learning layers for direct prediction?} While deep learning models (e.g., convolutional neural networks or recurrent neural networks) are well-suited for end-to-end classification, we opted to decouple feature extraction from classification for several reasons:
\begin{enumerate}
    \item \textbf{Interpretability:} Using a two-step process—where deep learning extracts high-level features and machine learning models handle classification—allows for more interpretable results. The use of SVM, kNN, and RF provides insights into how high-level features contribute to the classification decision.
    \item \textbf{Model Flexibility:} This approach gives flexibility to experiment with various machine learning classifiers on the same set of high-level features, making it easier to compare performance across different methods.
    \item \textbf{Performance Optimization:} Our experiments showed that deep learning models alone did not consistently outperform the hybrid approach. By utilizing pre-trained networks for feature extraction, we benefit from the high representational capacity of deep learning models while ensuring robust classification performance using traditional machine learning models.
    \item \textbf{Efficiency:} The separation of feature extraction and classification reduces computational overhead during classification, as simpler models such as SVM and RF are faster to train and deploy, particularly when reusing the deep-learned feature maps.
\end{enumerate}

\textbf{Comparison with Machine Learning-based Systems:} 
When comparing the high-level architectures of machine learning and deep learning-based systems, the primary difference lies in their approach to feature extraction. Machine learning-based systems typically extract hand-crafted features from both the time and frequency domains. In contrast, deep learning-based systems use deep neural networks to directly train normalized input data, bypassing the need for hand-crafted features. The resulting feature maps serve as the basis for deriving high-level features, which are then classified using the machine learning models.

\subsection{Proposed Deep Neural Networks for High-Level Feature Extraction}\label{dnn}

In this paper, we propose three types of deep neural network architectures for extracting high-level features.

\subsubsection{Supervised Deep Learning Model (SDLM)} 
\begin{figure}[h]
    \centering
    \includegraphics[width=1.0\linewidth]{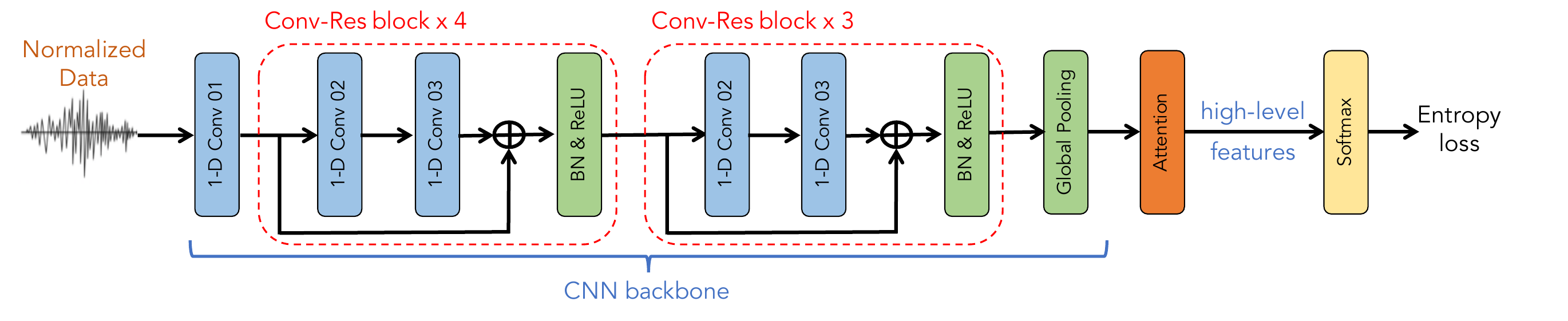}
    \caption{Architecture of the SDLM.}
    \label{fig:dl_01}
\end{figure}

The first deep neural network architecture is illustrated in Fig.\ref{fig:dl_01}.

As shown in Fig.\ref{fig:dl_01}, the proposed network architecture consists of four main components: one 1-dimensional convolutional block (1-D Conv 01), followed by two convolutional-residual blocks (Conv-Res), and finally, one attention block (Attention). The two convolutional-residual blocks share the same architecture, each of which includes two 1-D convolutional sub-blocks: 1-D Conv 02 and 1-D Conv 03, followed by batch normalization (BN)~\citep{batchnorm} and rectified linear unit (ReLU)~\citep{relu}. The detailed configuration of the 1-D Conv 01, 1-D Conv 02, and 1-D Conv 03 blocks is provided in Table~\ref{table:dl_01}.

\begin{table}[h]
    \caption{Configuration of Convolutional and Attention Blocks}
    \centering
    \begin{tabular}{cl} 
       \toprule
       \textbf{Blocks} & \textbf{Layers}  \\
       \midrule
       1-D Conv 01 & Conv[$48\times80$] - BN - ReLU - MP[$4$]\\
       1-D Conv 02 & Conv[$48\times3$] - BN - ReLU \\
       1-D Conv 03 & Conv[$96\times3$] - BN \\
       Attention  & Head=4, Key-Dimension=48\\
       \bottomrule
    \end{tabular}
    \label{table:dl_01} 
\end{table}

The output of the Conv-Res blocks is passed through a Global Average Pooling (GAP) layer, followed by an Attention layer \citep{vaswani2017attention}. The attention layer employs a Multihead attention scheme, empirically configured with 4 heads and a key dimension of 48, as illustrated in Table~\ref{table:dl_01}. Finally, a Softmax layer computes the predicted probabilities for specific classes.

As we utilize the Cross-Entropy loss, as shown in Equation (\ref{eq1}), to compare the predicted probabilities (i.e., the output of the Softmax layer) $\hat{\mathbf{y}}$ with the true labels $\mathbf{y}$.

\begin{equation}\label{eq1}
\mathcal{L}_{\text{entropy}} = -\sum_{i} y_i \cdot \log(\hat{y}_i)
\end{equation}

\subsubsection{Semi-Supervised Deep Learning Model (Semi-SDLM)} 
\begin{figure}[h]
    \centering
    \includegraphics[width=1.0\linewidth]{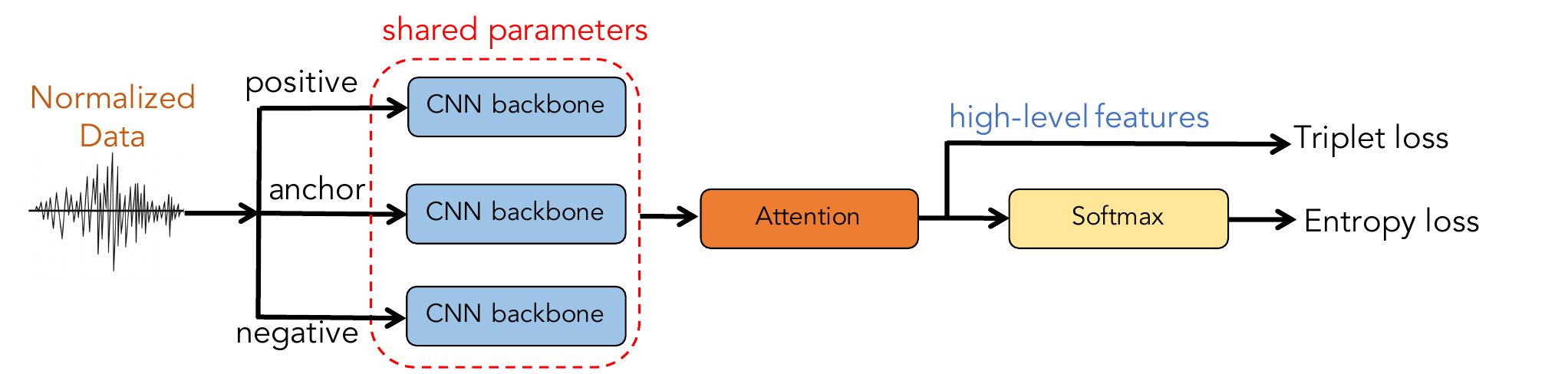}
    \caption{Architecture of the Semi-SDLM.}
    \label{fig:dl_02}
\end{figure}

This architecture adapts the SDLM by utilizing its initial layers up to the GAP layer as a convolutional neural network backbone (CNN backbone). The Semi-SDLM incorporates a triplet-based network, sharing this backbone, to learn from anchor, positive, and negative data inputs, illustrated in Fig.~\ref{fig:dl_02}.

The output feature maps from the CNN backbone serve as embeddings, which are subjected to two distinct loss functions. First, the Cross-Entropy loss (Equation (\ref{eq1})) classifies the embeddings into specific categories. Second, the Triplet loss~\citep{schroff2015facenet} is employed to maximize the distances between negative embeddings and groups of positive and anchor embeddings. For a batch of training samples indexed by $i$, the Triplet loss is formulated as:

\begin{equation}
    H_{i} = d^2(\mathbf{X}^{a}_{i}, \mathbf{X}^{p}_{i}) - d^2(\mathbf{X}^{a}_{i}, \mathbf{X}^{n}_{i}) + m, \notag
\end{equation}
\begin{equation}
    \mathcal{L}_{\text{triplet}} = \underset{\theta \in \mathbb{R}}{\text{argmin}}\sum_{i=1}^{n}\max( H_{i}, 0). \label{eq01}
\end{equation}

Here, $\mathbf{X}^{a}_{i}$, $\mathbf{X}^{p}_{i}$, and $\mathbf{X}^{n}_{i}$ denote embedding vectors from anchor, positive, and negative samples respectively, and $d^2(\mathbf{X}^{a}_i, \mathbf{X}^{p}_i)$ and $d^2(\mathbf{X}^{a}_i, \mathbf{X}^{n}_i)$ represent Euclidean distances between these embeddings. The margin $m$ enforces a minimum separation between positive and negative pairs. $n$ represents the number of triplets (anchor, positive, negative samples) considered in each training batch. The final loss function for training the Semi-SDLM, combining Cross-Entropy and Triplet losses, is:

\begin{equation}
    \mathcal{L}_{\text{Semi-SDLM}} = \mathcal{L}_{\text{triplet}} + \lambda \mathcal{L}_{\text{entropy}}
\end{equation}

where $\mathcal{L}_{\text{entropy}}$ is the Cross-Entropy loss, and $\lambda$ is a regularization parameter.

\subsubsection{Unsupervised Deep Learning Model (Unsupervised-DLM)} 
\begin{figure}[h]
    \centering
    \includegraphics[width=0.9\linewidth]{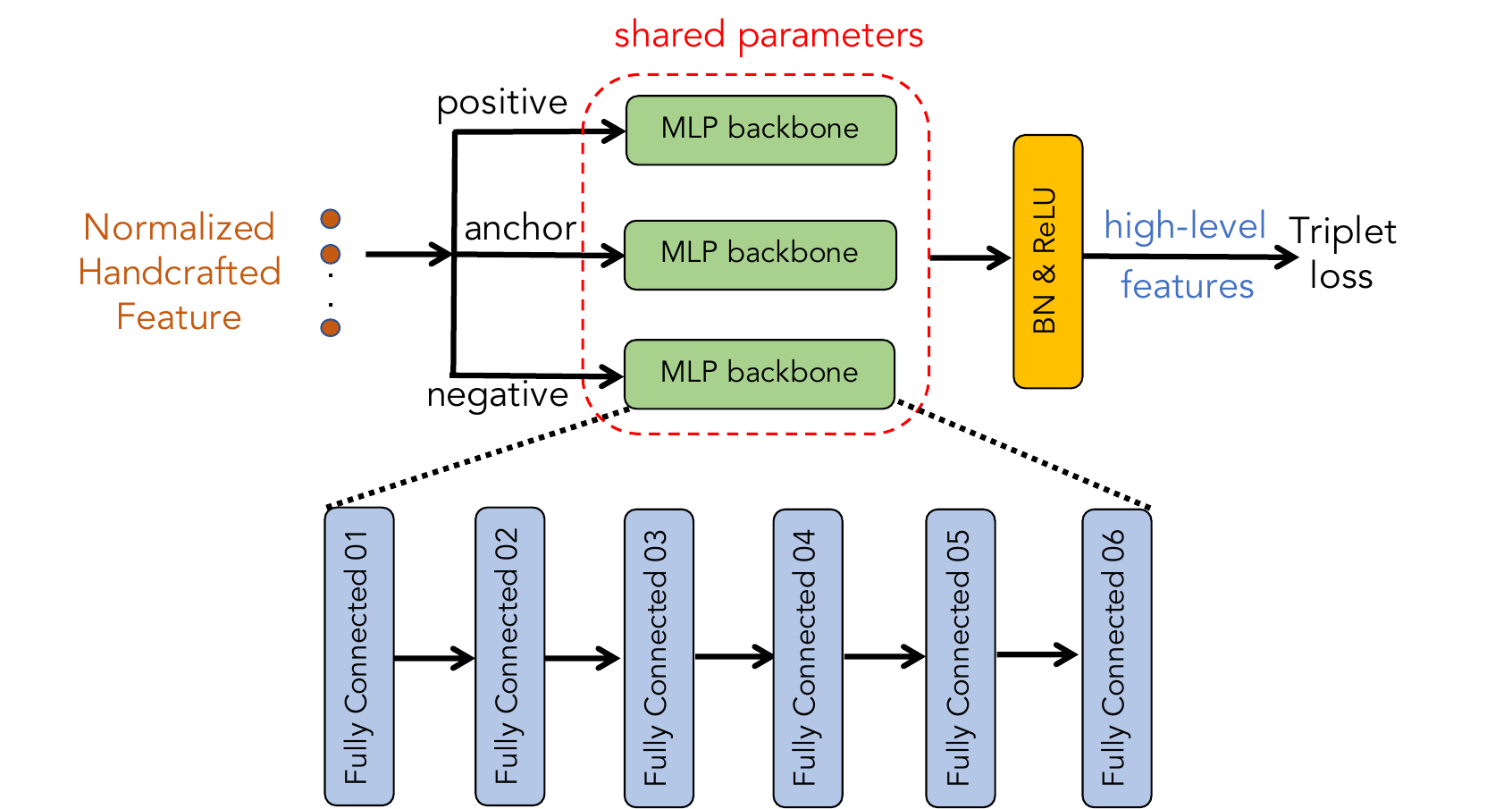}
    \caption{Architecture of the Unsupervised-DLM.}
    \label{fig:dl_03}
\end{figure}

This deep neural network employs a triplet network architecture, depicted in Fig.~\ref{fig:dl_03}. The input to this architecture consists of hand-crafted features used in the machine learning-based models discussed in Section~\ref{ml_sys}. However, the backbone of this network is a multilayer perceptron (MLP) structure, specifically a six-layer fully connected network with kernel sizes of 1024, 512, 128, 512, 1024, and 256. The architecture focuses solely on the Triplet loss function (Equation (\ref{eq01})) to learn feature representations from the output of the multilayer perceptron-based backbone (MLP backbone).

\subsubsection{High-Level Feature Extraction and Evaluation}\label{data_set}
For the SDLM and Semi-SDLM models, the high-level features are obtained from the output feature map of the Attention block. In contrast, for the Unsupervised-DLM model, the high-level features are derived from the output feature map of the MLP backbone. After training these three models, the extracted high-level features from the SDLM, Semi-SDLM, and Unsupervised-DLM models are used as input to traditional machine learning classifiers (e.g., SVM, RF, kNN) or distance-based methods (e.g., Cosine and Euclidean distances) for further classification, as illustrated in Fig.~\ref{fig:dl_fig}.

\section{Robust Machine Learning and Deep Learning-Based Fault Detection with a Novel Double Loss Function}

\begin{figure*}[th]
    \centering
    \includegraphics[width=1.0\linewidth]{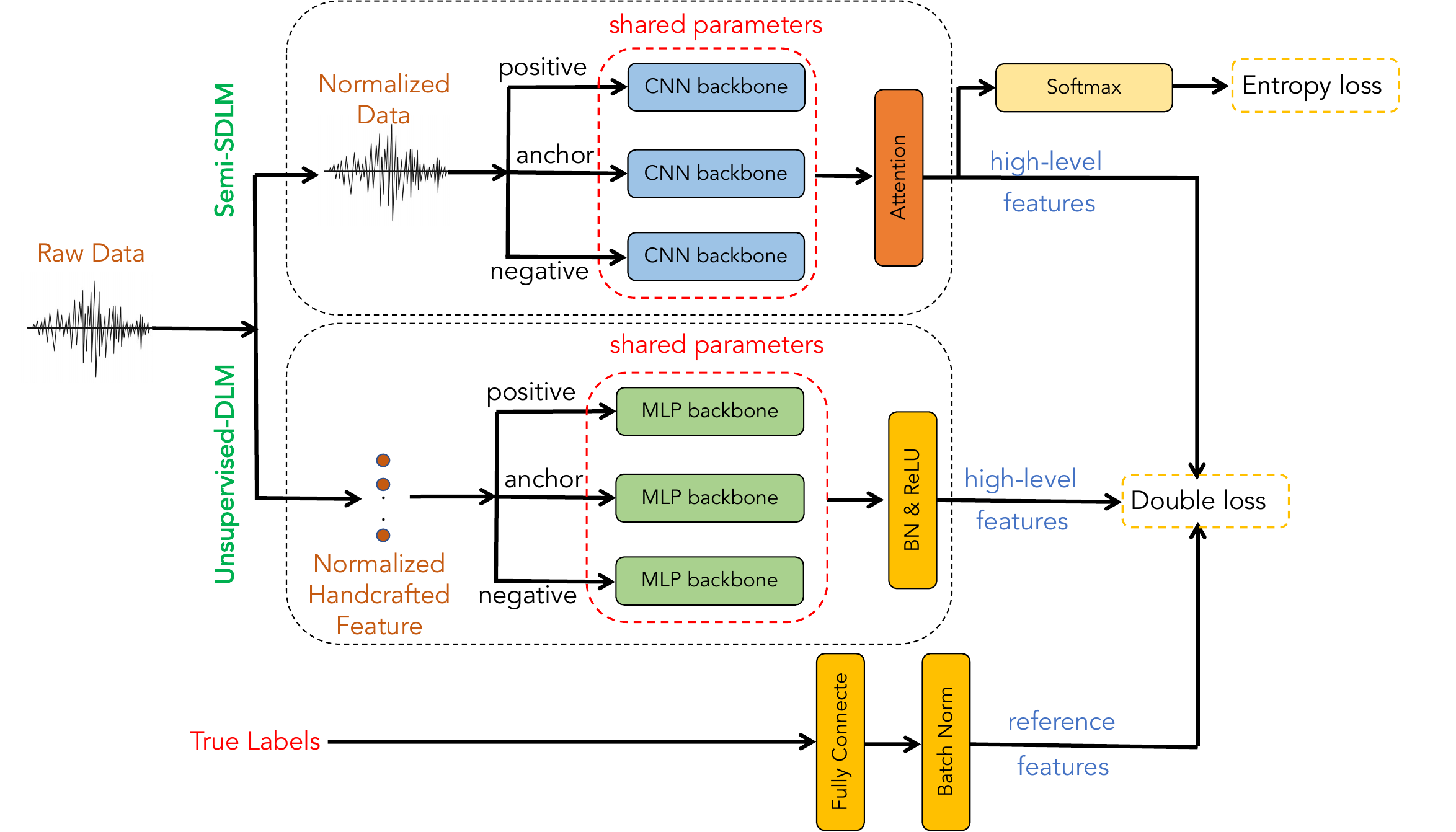}
    \caption{Architecture of the proposed robust deep learning-based system for bearing fault detection using the novel Double loss function (Robust-MBFD).}
    \label{fig_best}
\end{figure*}

To overcome the limitations of traditional Triplet losses, we introduce a novel Double loss function that combines Triplet loss and Center loss \cite{wen2016discriminative}. This approach defines consistent reference embedding vectors $\textbf{C}$, derived from true labels and processed through deep learning layers, as illustrated in Fig.~\ref{fig_best}. These reference vectors help minimize the distance between anchor embeddings $\textbf{X}^{a}$, enhancing the Center loss, represented as $d^2(\textbf{X}^{a}, \textbf{C})$.

The Center loss focuses solely on anchor embeddings, potentially neglecting positive and negative vectors. To address this, we integrate the Triplet loss, ensuring that all embedding vectors are appropriately positioned: vectors of the same class are close, while those from different classes are distant. This is expressed as:

\begin{equation}
    H_{i} = \gamma(d^2(\mathbf{X}^{a}_{i}, \mathbf{X}^{p}_{i}) + d^2(\mathbf{X}^{a}_{i}, \mathbf{C}_{i})) - d^2(\mathbf{X}^{a}_{i}, \mathbf{X}^{n}_{i}) + m.
\end{equation}

The parameter $\gamma$, empirically set to 0.4, balances the contributions of the loss components. The margin $m$ enforces separation between positive and negative pairs, while $H_{i}$ standardizes factors less than 0 for effective training.

The Double losses for the two branches of the Robust-MBFD model are defined as follows:

\begin{gather}
\begin{aligned}
    \mathcal{L}^{\text{Semi-SDML}}_{\text{double}} &= \underset{\theta \in \mathbb{R}}{\text{argmin}}\sum_{i=1}^{n}\max\left( H^{\text{Semi-SDML}}_{i}, 0\right), \\
    \mathcal{L}^{\text{Unsupervised-DML}}_{\text{double}} &= \underset{\theta \in \mathbb{R}}{\text{argmin}}\sum_{i=1}^{n}\max\left( H^{\text{Unsupervised-DML}}_{i}, 0\right).
\end{aligned}
\end{gather}

Combining these yields the overall loss function:

\begin{equation}
    \mathcal{L}_{2-Double} = \mathcal{L}^{\text{Semi-SDML}}_{\text{double}} + \mathcal{L}^{\text{Unsupervised-DML}}_{\text{double}}.
\end{equation}

The balance between these losses is regulated by parameter $\lambda$, empirically set to 0.3. Additionally, we combine this with Cross-Entropy loss \citep{ho2019real} for the first branch, resulting in the final loss for training the entire network:

\begin{equation}
    \mathcal{L} = \mathcal{L}_{entropy} + \Lambda \mathcal{L}_{2-Double},
\end{equation}

where $\Lambda$ is set to 0.01 to mitigate gradient perturbations due to the computational complexity of $\mathcal{L}_{2-Double}$.

This innovative approach enhances fault detection robustness by effectively utilizing both Triplet and Center losses, allowing for better class separation and improved model performance.

To evaluate our proposed models, we conducted experiments on three benchmark datasets of bearing vibration: The first dataset, PU Bearing \citep{pu_data}, comprises both lab-created and real damaged bearings. The other two datasets, CWRU Bearing \citep{cwru_data} and MFPT Bearing \citep{mfpt_data}, consist of artificially damaged bearings. The experimental settings, including the train/test split, some of them were initially based on guidelines provided in the dataset papers. However, to ensure the robustness of our proposed methods and mitigate potential biases, we further partitioned the data in a more challenging manner. This approach helps to validate the generalizability of our models across different scenarios and ensures that our training data is representative of diverse conditions encountered in practical applications.

\section{Experimental Settings} \label{Exp_Settings}

\subsection{PU Bearing Dataset}~\citep{pu_data} 
The bearing dataset, which was published by Paderborn University, currently presents the largest dataset of vibration data proposed for detecting damage in motors' bearings. There are a total of 32 different bearings used to collect vibration data in this dataset: 12 bearings with lab-created damages (i.e., These bearings, named KA01, KA03, KA05, KA06, KA07, KA08, and KA09, are for outer ring damages; The bearings KI01, KI03, KI05, KI07, and KI08 are marked for inner ring damages), 14 real damaged bearings collected from accelerated lifetime tests (i.e., The bearings KA04, KA15, KA16, KA22, KA30 are for outer ring damages; Inner ring damaged bearings are named KI04, KI14, KI16, KI17, KI18, KI21; The remaining bearings KB23, KB24, KB27 are used for both inner and outer ring damages), and 6 healthy bearings (i.e., These healthy bearings are referred to as K001, K002, K003, K004, K005, and K006 respectively).

Damage combination characterizes the occurrence of single, repetitive, or multiple damages in rolling bearings based on their symptoms: single damage affects one component (e.g., a single pitting on the inner ring), repetitive damage shows identical symptoms at several locations on the same component (e.g., multiple non-continuous pittings on the inner ring), and multiple damage involves different symptoms or identical symptoms on different components. The arrangement of these damages can be regular, random, or non-repetitive. Additionally, the geometrical size of damage is defined by its length, width, and depth according to VDI 3832 (2013). Damage characteristics are classified into single point and distributed damages based on whether only one rolling element is in contact with the damage, ensuring the damage is small relative to bearing size. The criteria are interdependent but provide a detailed damage description due to the diverse and mixed nature of real damages. The detailed information on bearing damages is provided in Table 1 of the study~\citep{pu_data}.

To evaluate this dataset, we follow the data split methods from the original paper \citep{pu_data}: (PU-C1) training on lab-created data and testing on real data (see Table~\ref{table:M1_PU}), addressing the challenge of limited labeled real data. Additionally, (PU-C2) involves training and testing on real data (see Table~\ref{table:M2_PU}), presenting a less challenging scenario. For PU-C2, as per the original methodology \citep{pu_data}, we conduct 10 training/testing combinations using three types of bearings for training and two for testing. The classification accuracy is averaged across these 10 combinations.

\begin{table}[h]
    \caption{Evaluation setup of PU Bearing dataset with training on lab-created data and testing on real data (PU-C1)} 
    \centering
    \begin{tabular}{cccc} 
       \toprule
     \textbf{Class} & \textbf{Fault type}  &\textbf{Training} &\textbf{Testing} \\
    \midrule
     1 &Healthy  &K002 &K001 \\
    \midrule 
     2 &OR Damage&KA01 &KA22 \\
       &         &KA05 &KA04 \\
       &         &KA07 &KA15 \\
       &         &     &KA30 \\
       &         &     &KA16 \\
     \midrule
     3 &IR Damage &KI01 &KI14 \\
       &          &KI05 &KI21 \\
       &          &KI07 &KI17 \\
       &          &     &KI18 \\
       &          &     &KI16 \\     
    \bottomrule
    \end{tabular}
    \label{table:M1_PU}  
\end{table}

\begin{table}[h]
    \caption{Evaluation setup of PU Bearing dataset with training and testing on real data (PU-C2)} 
    \centering
    \begin{tabular}{ccc} 
    \toprule
     \textbf{Healthy} &  \textbf{Outer Ring Damages} &\textbf{Inner Ring Damages} \\
    \midrule
     K001 &KA04 &KI04 \\
     K002 &KA15 &KI14 \\
     K003 &KA16 &KI16 \\
     K004 &KA22 &KI18 \\
     K005 &KA30 &KI21 \\                    
    \bottomrule
    \end{tabular}
    \label{table:M2_PU} 
\end{table}

Both splitting methods classify vibration samples into three categories: Healthy bearing, outer ring damaged bearing, or inner ring damaged bearing. It's important to note that each vibration sample in the PU Bearing dataset consists of an entire vibration recording file, encompassing approximately 255,900 data points (about 4 seconds).

\subsection{CWRU Bearing Dataset}~\citep{cwru_data}
The CWRU bearing vibration dataset, published by the Bearing Data Center at Case Western Reserve University, was recorded using a single type of motor under various operating conditions, including different bearing fault stages, motor speeds, load levels, and sensor positions on the bearing.

The dataset includes faults at different bearing locations (inner raceway, rolling element, and outer raceway) and varying fault diameters (0.007 inches to 0.040 inches). It also covers a range of motor speeds (1797 to 1720 RPM) and load levels, with sensor positions capturing faults in centered, orthogonal, or opposite positions on the bearing. For each unique operating condition, there is only one corresponding recording file. For instance, "IR007\_0" is the vibration recording file collected under the operating condition with a 0.007-inch fault diameter, a motor speed of 1797 RPM, and a fault at the inner race of the bearing.

Given the dataset's diverse range of operating conditions, we propose four different evaluation settings, as outlined in Table~\ref{table:M1_CWRU}. Specifically, we utilize all available healthy bearing data~\cite{cwru_data_normal}, which includes four healthy bearing data files: "Normal\_0," "Normal\_1," "Normal\_2," and "Normal\_3." For the fault bearing data, we focus exclusively on the drive end bearing type with a 12000 Hz sample rate~\citep{cwru_data_fault}, comprising 60 vibration recording files. In total, our analysis includes 64 vibration recording files.

\begin{table}[h]
    \caption{Overview of the CWRU datasets and the four proposed methods for training and testing splits.}
    \centering
    \begin{tabular}{cl} 
       \toprule
     \textbf{Cases} &\textbf{Class number and selected vibration data from CWRU dataset} \\
    \midrule
     CWRU-1  & Class 1 (Healthy): Normal\_0, Normal\_1, Normal\_2, Normal\_3; \\
        & Class 2 (Inner Race): IR007\_0, IR007\_1, IR007\_2, IR007\_3; \\
        & Class 3 (Ball): B007\_0, B007\_1, B007\_2, B007\_3; \\
        & Class 4 (Outer Race - Centered): OR007@6\_0, OR007@6\_1, OR007@6\_2, OR007@6\_3; \\
        & Class 5 (Outer Race - Orthogonal): OR007@3\_0, OR007@3\_1, OR007@3\_2, OR007@3\_3; \\
        & Class 6 (Outer Race - Opposite): OR007@12\_0, OR007@12\_1, OR007@12\_2, OR007@12\_3; \\

    \midrule 
    CWRU-2   & Class 1: Normal\_0   ; Class 2:  Normal\_1  ; Class 3:  Normal\_2  ; Class 4:  Normal\_3; \\
        & Class 5: IR007\_0    ; Class 6:  IR007\_1   ; Class 7:  IR007\_2   ; Class 8:   IR007\_3; \\   
        & Class 9: B007\_0     ; Class 10: B007\_1    ; Class 11: B007\_2    ; Class 12:  B007\_3; \\    
        & Class 13:OR007@6\_0  ; Class 14: OR007@6\_1 ; Class 15: OR007@6\_2 ; Class 16:  OR007@6\_3; \\ 
        & Class 17:OR007@3\_0  ; Class 18: OR007@3\_1 ; Class 19: OR007@3\_2 ; Class 20:  OR007@3\_3; \\ 
        & Class 21:OR007@12\_0 ; Class 22: OR007@12\_1; Class 23: OR007@12\_2; Class 24:  OR007@12\_3; \\
   \midrule	    
    CWRU-3   & Class 1: Normal\_0   ; Class 2:  Normal\_1  ; Class 3:  Normal\_2  ; Class 4:  Normal\_3; \\

        & Class 5: IR007\_0    ; Class 6:  IR007\_1   ; Class 7:  IR007\_2   ; Class 8:  IR007\_3; \\
        & Class 9: IR014\_0    ; Class 10: IR014\_1   ; Class 11: IR014\_2   ; Class 12: IR014\_3; \\   
        & Class 13: IR021\_0   ; Class 21: IR021\_1   ; Class 15: IR021\_2   ; Class 16: IR021\_3; \\   
        & Class 17: IR028\_0   ; Class 18: IR028\_1   ; Class 19: IR028\_2   ; Class 20: IR028\_3; \\   

        & Class 21: B007\_0     ; Class 22: B007\_1    ; Class 23: B007\_2    ; Class 24:  B007\_3; \\    
        & Class 25: B014\_0     ; Class 26: B014\_1    ; Class 27: B014\_2    ; Class 28:  B014\_3; \\    
        & Class 29: B021\_0     ; Class 30: B021\_1    ; Class 31: B021\_2    ; Class 32:  B021\_3; \\    
        & Class 33: B028\_0     ; Class 34: B028\_1    ; Class 35: B028\_2    ; Class 36:  B028\_3; \\    

        & Class 37:OR007@6\_0  ; Class 38: OR007@6\_1 ; Class 39: OR007@6\_2 ; Class 40:  OR007@6\_3; \\ 
        & Class 41:OR014@6\_0  ; Class 42: OR014@6\_1 ; Class 43: OR014@6\_2 ; Class 44:  OR014@6\_3; \\ 
        & Class 45:OR021@6\_0  ; Class 46: OR021@6\_1 ; Class 47: OR021@6\_2 ; Class 48:  OR021@6\_3; \\ 

        & Class 49:OR007@3\_0  ; Class 50: OR007@3\_1 ; Class 51: OR007@3\_2 ; Class 52:  OR007@3\_3; \\ 
        & Class 53:OR021@3\_0  ; Class 54: OR021@3\_1 ; Class 55: OR021@3\_2 ; Class 56:  OR021@3\_3; \\ 

        & Class 57:OR007@12\_0 ; Class 58: OR007@12\_1; Class 59: OR007@12\_2; Class 60:  OR007@12\_3; \\
        & Class 61:OR021@12\_0 ; Class 62: OR021@12\_1; Class 63: OR021@12\_2; Class 64:  OR021@12\_3; \\
   \midrule	    
    CWRU-4   & Class 1: Normal\_0   ,Normal\_1   ,Normal\_2   ,Normal\_3; \\

        & Class 2:  IR007\_0    ,IR007\_1    ,IR007\_2    ,IR007\_3, \\
        &           IR014\_0    ,IR014\_1    ,IR014\_2    ,IR014\_3, \\   
        &           IR021\_0   ,IR021\_1    ,IR021\_2    ,IR021\_3, \\   
        &           IR028\_0   ,IR028\_1    ,IR028\_2    ,IR028\_3; \\   

        & Class 3: B007\_0    ,B007\_1     ,B007\_2     ,B007\_3, \\    
        &          B014\_0    ,B014\_1     ,B014\_2     ,B014\_3, \\    
        &          B021\_0    ,B021\_1     ,B021\_2     ,B021\_3, \\    
        &          B028\_0    ,B028\_1     ,B028\_2     ,B028\_3; \\    
        
        & Class 4: OR007@6\_0,OR007@6\_1 ,OR007@6\_2,OR007@6\_3, \\ 
        &          OR014@6\_0,OR014@6\_1 ,OR014@6\_2,OR014@6\_3, \\ 
        &          OR021@6\_0,OR021@6\_1 ,OR021@6\_2,OR021@6\_3; \\ 
        &          OR007@3\_0,OR007@3\_1 ,OR007@3\_2,OR007@3\_3, \\ 
        &          OR021@3\_0,OR021@3\_1 ,OR021@3\_2,OR021@3\_3; \\ 
        &          OR007@12\_0,OR007@12\_1,OR007@12\_2,OR007@12\_3, \\ 
        &          OR021@12\_0,OR021@12\_1,OR021@12\_2,OR021@12\_3; \\
     \bottomrule
    \end{tabular}
    \label{table:M1_CWRU}
\end{table}

\subsection{MFPT Bearing Dataset}~\citep{mfpt_data} 
This vibration dataset, published by the Society for Machinery Failure Prevention Technology (MFPT), was recorded from a single type of bearing with consistent operating conditions (roller diameter: 0.235, pitch diameter: 1.245, number of elements: 8, and contact angle: zero). The dataset encompasses three types of bearing conditions: healthy bearings, fault bearings at the inner race, and fault bearings at the outer race.

The first category, healthy bearings, includes three vibration recording files, denoted as N1, N2, and N3, sample rate of 97,656 sps, for 5 seconds. These were recorded under a consistent load value of 270 lbs and an input shaft rate of 25 Hz.

The second category, fault bearings at the inner race, comprises seven vibration recordings (I1, I2, I3, I4, I5, I6, and I7), sample rate of 48,828 sps for 3 seconds, collected under varying load values: 0, 50, 100, 150, 200, 250, and 300 lbs, with an input shaft rate of 25 Hz.

The third category, fault bearings at the outer race, consists of ten vibration recordings. The first three (O1, O2, O3), sample rate of 97,656 sps for 6 seconds, were collected under a load setting of 270 lbs and an input shaft rate of 25 Hz. The remaining seven (O4, O5, O6, O7, O8, O9, O10), sample rate of 48,828 sps for 3 seconds, were recorded under load values of 25, 50, 100, 150, 200, 250, and 300 lbs, with an input shaft rate of 25 Hz. Additionally, the dataset includes real-world examples such as an intermediate shaft bearing from a wind turbine, an oil pump shaft bearing from a wind turbine, and a real-world planet bearing fault.

With the MFPT dataset, our objective is to classify vibration samples into one of three categories: healthy bearing, fault bearing at the inner race, or fault bearing at the outer race. Table~\ref{table:M1_MFPT} provides details of the official training and testing recordings following the dataset~\citep{mfpt_data}.

\begin{table}[h]
\caption{Official Training and Testing Sets for the MFPT Dataset}
\centering
\begin{tabular}{llll}
\toprule
\textbf{Type}             & \textbf{Healthy}   & \textbf{Outer Race}   & \textbf{Inner Race}  \\
\midrule
Training set  & $N_1, N_2$ & $O_{1}, O_{2}, O_{M1}$ & $I_1, I_2, I_4$\\
& & $ O_{M2}, O_{M4}, O_{M5}, O_{M7}$ & $I_5, I_7$\\
\midrule
Test set      & $N_2, N_3$ & $O_3, O_{M3}, O_{M6}$ & $I_3, I_6$\\
\bottomrule
\end{tabular}
\label{table:M1_MFPT} 
\end{table}

For each vibration recording file, we segment the entire recording into short-time segments of 400 ms with a 50\% overlap, referred to as vibration samples. These samples are then used as input for our proposed classification models. In essence, our models operate on short-time segments of 400 ms.

\subsection{Evaluation Metric}\label{metric}
We follow the recommendation in \cite{pu_data}, the paper introducing the PU Bearing dataset, using classification accuracy as the primary metric for model evaluation. Let $M$ denote the number of correctly predicted vibration samples out of total samples $N$. The classification accuracy (Acc.\%) is calculated as:

\begin{equation}
    \text{Acc.\%} = 100 \times \frac{M}{N}
\end{equation}

\subsection{Model Implementation}\label{model}
We implement traditional machine learning models (SVM, DT, RF) using the Scikit-learn toolkit~\citep{Scikit}.
Meanwhile, the proposed deep learning models are built on the TensorFlow framework.
All deep learning models are optimized using the Adam algorithm~\citep{Adam}.
The training and evaluation processes are conducted on a GPU Titan RTX with 24GB of memory.
The training processes of deep learning models stop after 100 epochs. A batch size of 16 is used throughout the entire process.

\section{Experimental Results and Discussion}
\label{result}

Among the benchmark datasets discussed in Section~\ref{data_set}, the PU dataset \citep{pu_data} includes both lab-created and real data. The PU-C1 splitting method involves training on lab-created data and evaluating on real data, addressing the challenge of limited real-world data availability. Thus, we initially evaluate our proposed models using the PU dataset \citep{pu_data} and the PU-C1 splitting method. Hyperparameters and settings for both our machine learning models and proposed deep learning architectures were determined through extensive experimentation.

\subsection{Evaluation on the PU Dataset Using the PU-C1 Splitting Method: Achieving the Best-Proposed Method with a Novel Loss Function}

\subsubsection{Machine Learning-Based Systems Evaluation} 

In particular, we evaluate a wide range of machine learning models, specifically SVM, kNN, and RF, configured as shown in Table~\ref{table:set_model}. We utilize six different normalization methods: MaxAbsScaler (MAS)~\citep{ichimura2019route}, StandardScaler (SS)~\citep{bisong2019introduction}, RobustScaler (RS)~\citep{dhali2020efficient}, Normalizer (N)~\citep{freitas2023ripening}, QuantileTransformer (QT)~\citep{takkala2022kyphosis}, and PowerTransformers (PT)~\citep{velasquez2021support}, implemented using the Scikit-Learn toolbox~\citep{pedregosa2011scikit}. Additionally, we consider three types of features: only time-domain features, only frequency-domain features, and a combination of both time and frequency-domain features. The results of these machine learning models are presented in Table~\ref{table:res_01}. As shown in Table~\ref{table:res_01}, the kNN model with the Normalizer method achieves the highest accuracy of 50.45\%. The kNN model also yields the best results when combined with other normalization methods and different types of features. Furthermore, with the QuantileTransformer and PowerTransformer normalization methods, the RF and SVM models achieve their highest accuracy scores of 47.95\% and 45.45\%, respectively, when using only time-domain features.

\begin{table*}[h]
    \caption{Performance comparison (Acc \% ) among machine learning-based models on the PU dataset~\citep{pu_data} using the PU-C1 splitting method (training on lab-created data and evaluating on real data).}
    \centering
    \begin{tabular}{cccccccccc} 
    \toprule
            & \multicolumn{3}{c}{\textbf{Time}} & \multicolumn{3}{c}{\textbf{Frequency}} &  \multicolumn{3}{c}{\textbf{Time \& Frequency}} \\
    \midrule
      &\textbf{SVM} &\textbf{kNN} &\textbf{RF} &\textbf{SVM} &\textbf{kNN} &\textbf{RF}  &\textbf{SVM} &\textbf{kNN} &\textbf{RF}  \\
    \midrule
     MAS        & 43.41 & 49.20          & 46.70  & 43.97 & 50.22          & 43.97 & 43.97 & \textbf{50.34} & 44.09 \\
     SS      & 39.43 & \textbf{49.43} & 45.90  & 44.09 & 48.63          & 43.97 & 44.09 & 48.86          & 43.75 \\
     RS        & 46.81 & 47.04          & 47.38  & 42.84 & \textbf{49.09} & 44.09 & 42.84 & \textbf{49.09} & 43.97 \\
     N          & 42.72 & 43.52          & 34.31  & 41.25 & \textbf{50.45} & 43.97 & 48.75 & 41.47          & 30.22 \\
	QT  & 45.00 & 40.91          & \textbf{47.95}  & 43.97 & 43.63          & 43.63 & 43.97 & 43.40          & 43.97 \\
	   PT     & \textbf{45.45} & 44.09          & 44.65  & 44.09 & 43.52          & 43.75 & 44.09 & 43.29          & 43.75 \\
    \bottomrule
    \end{tabular}
    \label{table:res_01} 
\end{table*}

\subsubsection{SDLM Evaluation} We then evaluate the SDLM. This model is also evaluated with six normalization methods (e.g., MaxAbsScaler, StandardScaler, RobustScaler, Normalizer, QuantileTransformer, and PowerTransformer) and a wide range of back-end classification models, including SVM, RF, kNN, Euclidean, and Cosine. As Table~\ref{table:res_02} shows, the three best scores of 53.06, 56.70, and 56.70 are achieved with the combinations of Normalizer and SVM, RobustScaler and Euclidean, and RobustScaler and Cosine, respectively. These results also indicate that the supervised deep learning-based systems outperform the machine learning-based systems (i.e., the best performance of a machine learning-based model only reaches 50.45 with Normalizer and kNN).

\begin{table}[h]
    \caption{Performance comparison (Acc \%) of the SDLM system on the PU dataset~\citep{pu_data} using the PU-C1 splitting method (training on lab-created data, testing on real data).}

    \centering
    \begin{tabular}{ccccccc}
    \toprule
      & \textbf{MAS} & \textbf{SS} & \textbf{RS} & \textbf{N} & \textbf{QT} & \textbf{PT}\\
    \midrule
     SVM       & 44.88 & 44.31 & 44.54 & \textbf{53.06} &  47.72 & 43.86\\
     RF        & 46.93 & 47.95 & 45.22 & 47.04 &  47.15 & 47.72 \\
     kNN       & 45.45 & 45.45 & 45.45 & 38.97 &  45.45 & 45.45 \\
     Euclidean & 34.20 & 42.15 & \textbf{56.70} & 32.27 &  52.15 & 34.43\\
     Cosine    & 34.20 & 42.15 & \textbf{56.70} & 32.27 &  52.15 & 34.43\\
    \bottomrule
    \end{tabular}
    \label{table:res_02} 
\end{table}

\subsubsection{Semi-SDLM and Unsupervised-DLM Evaluation}
Given the promising results of the SDLM, which outperformed the machine learning-based systems, we further evaluate the Semi-SDLM and Unsupervised-DLM models proposed in Section~\ref{dnn}. In this experiment, we use a single normalization method (e.g., Normalizer) to compare the performance of these deep learning architectures. Table~\ref{table:res_03} shows that SDLM, Semi-SDLM, and Unsupervised-DLM achieve the same best score of 53.06\% with the SVM classifier.

\begin{table}[h]
    \caption{Performance comparison (Acc \%) of deep learning-based models on the PU dataset~\citep{pu_data} using the PU-C1 splitting method. Models include Normalizer for data normalization, and SDLM, Semi-SDLM, and Unsupervised-DLM for high-level feature extraction, with SVM for back-end classification.}
    \centering
    \begin{tabular}{cccc} 
    \toprule
     \textbf{Models} & \textbf{SDLM} & \textbf{Semi-SDLM} & \textbf{Unsupervised-DLM} \\
    \midrule
     SVM       & \textbf{53.06} & \textbf{53.06} & \textbf{53.06} \\
     RF        & 48.75 & 47.04 & 49.43 \\
     KNN       & 38.97 & 38.97 & 38.97 \\
     Euclidean & 32.27 & 32.27 & 49.54 \\
     Cosine    & 32.27 & 32.27 & 46.81 \\
    \bottomrule
    \end{tabular}
    \label{table:res_03} 
\end{table}

\subsubsection{Achieving the Best Performance: Robust-MBFD with a Novel Double Loss Function}
Given that each of the deep learning networks (e.g., SDLM, Semi-SDLM, Unsupervised-DLM) extracts distinct high-level features due to their different architectures and training strategies, there is significant potential to enhance the performance of the entire MBFD system by combining these networks to generate high-performing high-level features. We propose a novel deep learning-based system for bearing fault detection, which integrates the training strategies and network architectures of SDLM, Semi-SDLM, and Unsupervised-DLM. We refer to this integrated model as Robust-MBFD. The proposed Robust-MBFD system is illustrated in Fig.~\ref{fig_best}. As shown in the figure, Robust-MBFD comprises two branches. In the first branch, the normalized raw data is fed into a CNN backbone, which is adapted from the SDLM architecture (see Figs.~\ref{fig:dl_01} and~\ref{fig:dl_02}). In the second branch, handcrafted features (e.g., time and frequency domain features mentioned in Section~\ref{ml_sys}) are extracted from the raw data. These handcrafted features are normalized before being input into a MLP backbone, which is adapted from the Unsupervised-DLM model (see Fig.~\ref{fig:dl_03}). For the output of the CNN backbone in the first branch, we apply two loss functions: Entropy loss for supervised learning and Triplet loss for unsupervised learning.


This approach is similar to the use of two loss functions mentioned in the Semi-SDML model. For the output of the MLP backbone in the second branch, we apply the Triplet loss, similar to the Unsupervised-SDML model. By using the Triplet loss on both outputs from the CNN backbone and the MLP backbone, embedding vectors within the same class are brought closer, while those from different classes are pushed farther apart. However, grouping inconsistent embedding vectors together, as suggested by the Triplet loss study \citep{schroff2015facenet}, can lead to disadvantages: (1) Overfitting can occur in several separate groups instead of one group in a class; (2) Changing weights in models for grouping vectors can be time-consuming if there is no consistent vector used as a reference for others entering the same class.

The Robust-MBFD system, enhanced with the novel Double loss, achieved a significant accuracy score of 72.72\%, as demonstrated in Table \ref{res_rob}. This marks an improvement of approximately 20.00\% compared to individual deep learning base systems such as SDLM, Semi-SDLM, or Unsupervised-DLM, all utilizing the same Normalizer for normalization and SVM for back-end classification, as detailed in Table \ref{table:res_03}.

These results strongly underscore the effectiveness of the Robust-MBFD system with the Double loss for bearing fault detection. Comparative experiments against the Triplet loss and Center loss, using the network architecture depicted in Fig.~\ref{fig_best}, confirm the superior performance of the Double loss, as depicted in Table~\ref{res_rob}. By integrating the Robust-MBFD system with the Double loss, Normalizer, and employing the Euclidean measurement in the backend, we achieved the highest accuracy of 72.72\% on the PU dataset with PU-C1 splitting.

\begin{table}[h]
\caption{Performance comparison (Acc \%) of the Robust-MBFD system using Double loss with Triplet loss and Center loss, evaluated on the PU dataset~\citep{pu_data} using the PU-C1 splitting method (training on lab-created data and evaluating on real data).}

\centering
\begin{tabular}{llll}
\toprule
\textbf{Method} & \textbf{Triplet loss} & \textbf{Center loss} & \textbf{Double loss} \\
\midrule
SVM & 55.68 & 51.36 & 65.45 \\
RF & 52.61 & 51.70 & 63.29 \\

KNN & 42.27 & 49.20 & 61.81 \\
Euclidean & 42.50 & 49.43 & \textbf{72.72} \\
Cosine & 43.29 & 48.75 & 32.72 \\
Ensemble & 48.18 & 49.65 & 65.22 \\
\bottomrule
\end{tabular}
\label{res_rob}
\end{table}

The table \ref{table:sizeOfProposed} compares the size of our proposed Robust-MBFD model with various versions of MobileNet \citep{howard2019searching}, a leading deep learning model for low-cost devices. Our model exhibits competitive size characteristics, akin to state-of-the-art models. Additionally, to optimize for embedded systems, we will convert our model to TensorFlow Lite, reducing its size for deployment on resource-constrained hardware.

\begin{table}[h]
\centering
\caption{Size Comparison of the Robust-MBFD System with Various Versions of MobileNet}\label{table:sizeOfProposed} 
\vspace{0.3cm}
\begin{tabular}{lll}
\toprule
\textbf{Model}       & \textbf{Params}  & \textbf{Size (MB)} \\
\midrule
Our Robust-MBFD & 11,468,800 & 4.4 \\
mobilenet\_v3\_large\_1.0\_224  & 13,926,400 & 5.4 \\
mobilenet\_v3\_large\_0.75\_224 & 10,485,760 & 4.0 \\
mobilenet\_v3\_large\_minimalistic\_1.0\_224 & 10,158,080 & 3.9 \\
mobilenet\_v3\_small\_1.0\_224 & 7,549,740 & 2.9 \\
mobilenet\_v3\_small\_0.75\_224 & 6,291,456 & 2.4 \\
mobilenet\_v3\_small\_minimalistic\_1.0\_224 & 5,242,880 & 2.0 \\
\bottomrule
\end{tabular}
\end{table}

\subsection{Evaluation of the Proposed Robust-MBDF System on Various Datasets}
\label{5a}

As the proposed Robust-MBFD and the novel Double loss function help to achieve the best performance on the PU dataset with the splitting method of PU-C1, we further evaluate this system with the PU dataset using the PU-C2 splitting method and the other datasets of CWRU Bearing and MFPT Bearing using the splitting methods mentioned in Section~\ref{data_set}.

\subsubsection{PU Dataset with PU-C2 Splitting} 
The results from Table~\ref{table:res_04} demonstrate that our proposed robust deep learning-based system achieves superior performance on the PU dataset using the PU-C2 splitting method. Specifically, it achieves the highest accuracies of 90.91\% and 91.41\% when employing PowerTransformer normalization combined with Euclidean and Cosine distance measurements, respectively. These findings underscore the effectiveness of our approach in leveraging advanced normalization techniques and distance measures to enhance classification accuracy on complex datasets such as PU. These results highlight that Euclidean and Cosine distance measurements significantly outperform traditional machine learning models (SVM, RF, kNN) in accurately classifying concatenated high-level features extracted from vibration data. This improvement signifies the robustness and scalability of our proposed deep learning-based system for motor bearing fault detection applications.

\begin{table}[h]
\centering
\caption{Evaluation of the proposed Robust-MBFD system on the PU dataset using the PU-C2 splitting method (training and evaluating on real data).}\label{table:res_04} 
\vspace{0.3cm}
\begin{tabular}{@{}lllllll@{}}
\toprule
         & \textbf{MAS}   & \textbf{SS}       & \textbf{RS}    & \textbf{N}   & \textbf{QT} & \textbf{PT}\\
\midrule
SVM       & 78.64 & 73.30  & 78.60 & 68.67 & 79.35 & 89.25\\
RF        & 78.32 & 69.50  & 73.50 & 74.40 & 88.36 & 88.75\\
KNN       & 68.25 & 76.60  & 78.90 & 78.40 & 78.70 & 84.08\\
Euclidean & 73.36 & 74.40  & 77.63 & 77.61 & \textbf{88.60} & \textbf{90.91}\\
Cosine    & 79.10  & 79.43 & 73.10 & 77.23 & 88.23 & \textbf{91.41}\\
\bottomrule
\end{tabular}
\end{table}

The high performance on the PU dataset with the PU-C2 splitting method, compared to the lower performance with the PU-C1 splitting method, is attributed to the use of only real-world data for both training and evaluation in PU-C2. In contrast, PU-C1 uses lab-created data for training and real-world data for validation. This indicates the potential applicability of our approach in real-life scenarios when large amounts of authentic data are available.

\subsubsection{CWRU Dataset} 
The experimental results on the CWRU dataset with four splitting methods (e.g., CWRU-C1, CWRU-C2, CWRU-C3, CWRU-C4) are presented in Table~\ref{table:res_05}. As shown in Table~\ref{table:res_05}, almost all high accuracy scores are obtained from the Euclidean and Cosine distance measurements with PowerTransformer normalization. These results are similar to those observed on the PU dataset with the PU-C2 splitting method.

\begin{table}[h!]
\centering
\caption{Evaluation of the proposed Robust-MBFD system on the CWRU datasets}\label{table:res_05} 
\begin{tabular}{@{}lllllll@{}}
\toprule
            & \textbf{MAS}   & \textbf{SS}       & \textbf{RS}    & \textbf{N}   & \textbf{QT} & \textbf{PTF}\\
\midrule
CWRU-C1  & & & & & & \\
\midrule
SVM    & 38.71 & 69.38 & 68.42 & 83.27 & 84.05 & 69.55\\
RF     & 55.53 & 66.07 & 67.11 & 62.89 & 69.03 & 66.85\\
KNN    & 60.67 & 52.70 & 52.17 & 74.52 & 55.09 & 52.48\\
Euclidean & 63.37 & 78.57 & 88.93 & 71.90 & 85.97 & \textbf{89.15}\\
Cosine & 62.15 & 78.91 & \textbf{88.89} & 69.72 & 85.27 & \textbf{87.71}\\
\midrule
CWRU-C2  & & & & & & \\
\midrule
SVM    & 42.76 & 46.32 & 63.54 & 62.30 & 70.06 & 65.23\\
RF     & 41.89 & 47.77 & 47.45 & 39.36 & 47.84 & 48.26\\
kNN    & 41.85 & 41.18 & 41.00 & 49.13 & 39.02 & 41.34 \\
Euclidean & 35.78 & 69.94 & 69.92 & 56.57 & 70.48 & \textbf{89.63}\\
Cosine & 34.23 & 68.54 & 70.02 & 57.63 & \textbf{71.56} & \textbf{90.01}\\
\midrule
CWRU-C3  & & & & & & \\
\midrule
SVM    & 25.30 & 45.23 & 29.56 & 55.57 & 56.68 & 29.65\\
Rf     & 33.89 & 38.17 & 38.43 & 28.33 & 38.28 & 38.65 \\
kNN    & 36.96 & 35.19 & 35.60 & 43.54 & 29.18 & 35.34\\
Euclic & 57.10 & 79.66 & 78.73 & 51.80 & 61.95 & \textbf{80.54} \\
Cosine & 58.63 & \textbf{81.78} & 77.21 & 52.23 & 60.89 & \textbf{81.57}  \\
\midrule
CWRU-C4  & & & & & & \\
\midrule
SVM    & 77.03 & 88.39 & 82.41 & 83.88 & 87.15 & 90.32\\
RF     & 63.68 & 68.42 & 69.08 & 64.33 & 68.32 & 68.52 \\
kNN    & 66.66 & 61.45 & 61.22 & 79.99 & 55.06 & 61.59\\
Euclidean & 72.89 & \textbf{91.65} & \textbf{90.44} & 78.73 & 83.51 & \textbf{90.35}\\
Cosine & 73.68 & 90.32 & 89.66 & 77.21 & 84.69 & 89.65\\
\bottomrule
\end{tabular}
\end{table}

For all four splitting methods, the proposed robust deep learning-based system consistently achieves accuracy rates of over 90.0\%. This demonstrates the robustness of the proposed system on the CWRU dataset.

\subsubsection{MFPT Dataset}
Looking at Table~\ref{table10}, we achieved very high performance on the MFPT dataset with the best accuracy score of 99.9\% (SVM for the back-end classification and SS for the normalization). Although high accuracy scores are observed in machine learning-based classification methods such as SVM (99.9\% with SS normalization and 99.8\% with RS normalization) or RF (99.8\% with SS normalization), the results from Euclidean and Cosine measurements are very competitive, with the best scores being 99.7\% and 99.6\% with SS normalization. We also observe that scores with SS normalization consistently achieve over 99.6\% in this dataset.

\begin{table}[h]
\centering
\caption{Evaluation of the proposed Robust-MBFD system on the MFPT Dataset}
\begin{tabular}{@{}lllllll@{}}
\toprule
         & \textbf{MAS}   & \textbf{SS}   & \textbf{RS}    &\textbf{N}   & \textbf{QT} & \textbf{PTF}\\
\midrule
SVM    & 99.11 & \textbf{99.90}  & \textbf{99.86} & 89.22 & 98.78  & 97.51\\
RF     & 97.63 & \textbf{99.79} & 99.73 & 86.82 & 98.35  & 99.09\\
kNN    & 98.04 & 99.76 & 99.80  & 87.11 & 96.15  & 96.82\\
Euclic & 98.70 & 99.66 & 99.56 & 89.30 & 98.32 & 99.12 \\
Cosine & 99.06 & 99.72 & 99.32 & 89.70 & 98.48 & 99.28 \\
\bottomrule
\end{tabular}
\label{table10}
\end{table} 

We acknowledge that the state-of-the-art methods for motor bearing fault detection exhibit certain limitations, particularly in terms of input and output parameter settings. Many existing approaches lack a comprehensive analysis of how variations in these parameters, such as bearing fault stage, motor speed, load level, and drive type, impact classification accuracy. Our study addresses these gaps by incorporating a detailed examination of these parameters within our experimental settings. Each fault state and operational condition was meticulously analyzed to understand their distinct impact on the system’s performance. Specifically, with the CWRU and MFPT datasets, which present various challenging settings including motor speed, load level, and drive type as shown in Section \ref{Exp_Settings}, we achieved high accuracy scores around $90\%$, as detailed in Tables \ref{table:res_05} and \ref{table10}. However, with different fault stages as shown in Section \ref{Exp_Settings}, and when training on lab-created data and validating on real-world data in the PU dataset, the accuracy score dropped to around $72.72\%$, as shown in Table \ref{res_rob}. This highlights the significant impact that variations in these parameters can have on classification accuracy. Our study provides a more thorough understanding of the operational dynamics that influence fault detection accuracy. This comprehensive analysis helps in setting more reliable benchmarks and paves the way for future research to build upon a more detailed and parameter-sensitive foundation.

\section{Conclusion}
\label{conclusion}
In this paper, we introduced Robust-MBFD, a robust deep learning-based system specifically designed for motor bearing fault detection (MBFD). Our approach integrates three distinct deep neural network architectures that employ supervised, semi-supervised, and unsupervised learning strategies, augmented by a novel Double loss function. This innovative combination aims to extract concise and distinctive high-level features from raw vibration data, enabling effective detection of motor bearing faults using traditional machine learning models (e.g., SVM, kNN, RF) and popular distance metrics (e.g., Euclidean, Cosine). Extensive experiments conducted on three benchmark datasets—PU Bearing, CWRU Bearing, and MFPT Bearing—using various data splitting strategies, demonstrate the generalizability and robustness of the proposed Robust-MBFD system. The results highlight its superior efficacy in real-world motor bearing fault detection applications. Additionally, the proposed system and our suggested data splitting methods for these datasets provide valuable benchmarks and settings for future research in the field of MBFD. The integration of diverse training strategies and the innovative Double loss function significantly enhances the performance of the Robust-MBFD system. This work not only demonstrates the potential of deep learning approaches in MBFD but also sets a new standard for robustness and accuracy in fault detection systems.

\section*{Contact Information}
For access to the code and further information about these proposed systems, please contact AIWARE Limited Company at:\\
\url{https://aiware.website/Contact}

\section*{Preprint Availability}
A preprint of this manuscript is available at:\\
\url{https://arxiv.org/abs/2310.11477}

\section*{Acknowledgment}
This research is funded by the Vietnamese Ministry of Education and Training under project number B2024.DNA.18

\bibliography{sn-bibliography}

\end{document}